\documentclass[letterpaper]{article} 
\usepackage{aaai2026}  
\usepackage{times}  
\usepackage{helvet}  
\usepackage{courier}  
\usepackage[hyphens]{url}  
\usepackage{graphicx} 
\usepackage{natbib}  
\usepackage{caption} 
\usepackage{algorithm}
\usepackage{algorithmic}

\usepackage{newfloat}
\usepackage{listings}

\DeclareCaptionStyle{ruled}{labelfont=normalfont,labelsep=colon,strut=off} 
\floatstyle{ruled}
\newfloat{listing}{tb}{lst}{}
\floatname{listing}{Listing}
\usepackage[table]{xcolor} 
\usepackage{booktabs}
\usepackage{amsmath} 
\usepackage{tcolorbox} 
\usepackage{cancel} 
\usepackage{multirow} 
\usepackage{makecell} 

\title{SCB-Dataset: A Dataset for Detecting Student and Teacher Classroom Behavior}
\author{
    Fan Yang
}
\affiliations{
    \textsuperscript{\rm }Jinan University, Guangzhou, China, winstonyf@qq.com\\

}

\usepackage{bibentry}

\begin{document}

\maketitle

\begin{abstract}
Using deep learning methods to detect the classroom behaviors of both students and teachers is an effective way to automatically analyze classroom performance and enhance teaching effectiveness. Then, there is still a scarcity of publicly available high-quality datasets on student-teacher behaviors. We constructed SCB-Dataset—a comprehensive dataset of student and teacher classroom behaviors covering 19 classes. SCB-Dataset is divided into two types: Object Detection and Image Classification. The Object Detection part includes 13,330 images and 122,977 labels, and the Image Classification part includes 21,019 images. We conducted benchmark tests on SCB-Dataset using YOLO series algorithms and Large vision-language model. We believe that SCB-Dataset can provide a solid foundation for future applications of artificial intelligence in education.

\end{abstract}
\begin{links}
    \link{Code}{https://github.com/Whiffe/SCB-dataset}
\end{links}


\section{Introduction}
The rapid development of artificial intelligence (AI), especially deep learning, has led to significant development in the field of computer vision, particularly the rapid development of Large vision-language models (LVLMs) in the recent two years \citep{jaech2024openai, team2024gemini, bai2025qwen2}, which has brought subversive changes to the entire industry, such as object recognition \citep{wang2015action, wang2018non, wang2017spatiotemporal}, object detection \citep{liu2016ssd, redmon2016you, ren2015faster, chen2023diffusiondet,zong2023detrs}, object tracking \citep{berclaz2006robust, breitenstein2009robust, defferrard2016convolutional}, instance segmentation \citep{He_2017, woo2023convnext}, video retrieval \citep{ma2015multimodal, wang2015learning, wang2017survey}, visual question answering (VQA) \citep{ma2016learning}, scene understanding, and visual reasoning \citep{bai2025qwen2}, Video-based Action Recognition \citep{tran2015learning, wang2023videomae, feichtenhofer2019slowfast}, etc. Algorithms generally perform well on simple datasets, such as the COCO \citep{lin2014microsoft}, Crowded Human \citep{shao2018crowdhuman}, UCF101 \citep{soomro2012ucf101}, HMDB \citep{kuehne2011hmdb}, and MSVD \citep{chen2011collecting} datasets. However, due to the complexity and diversity of real life, models that perform well on simple datasets cannot meet real-world needs. Currently, remarkable progress has been made in constructing increasingly complex and realistic datasets, such as the AVA \citep{gu2018ava} and VATEX \cite{wang2019vatex} datasets. However, there is still a lack of public high-quality complex and realistic datasets in the education field, which greatly limits the development of artificial intelligence in this field. Based on these findings, this study establishes a large-scale public dataset for the field of education.

Evaluations of education quality have attracted an increasing amount of attention from researchers in fields such as pedagogy and psychology.  As a basic teaching form, classroom teaching has always been the core of education.  As part of a certain scenario, students’ and teachers' behaviors in a classroom are significant and not disregarded. Acquiring information about student and teacher behaviors is not only helpful for mastering students' learning, personality, and psychological traits, providing feedback on problems in teachers' teaching processes, but also worthy of inclusion in evaluations of education quality.

\begin{figure}[t]
\centerline{\includegraphics[width=0.5\textwidth]{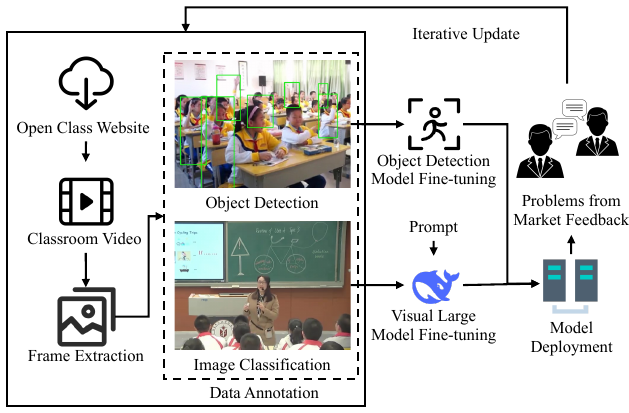}}
\caption{The production process of SCB-Dataset.}
\label{SCB-Process}
\end{figure}

\begin{figure*}[t]
\centerline{
\includegraphics[width=0.88\textwidth]{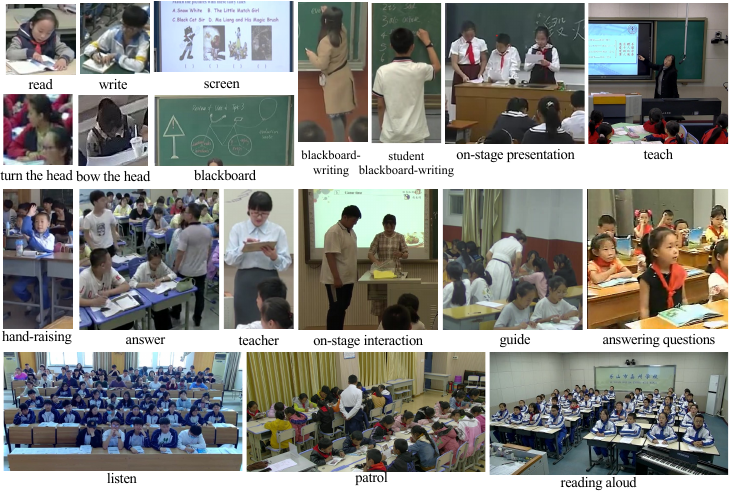}
}
\caption{Examples of behavior classes in SCB-Dataset}
\label{SCB_samples}
\end{figure*}

With the advent of the era of big data, there are a large number of high-quality classroom videos publicly available on video websites, which provides a large amount of raw data for the dataset production of this paper. The data production process of this paper is shown in Fig.~\ref{SCB-Process}. First, download classroom videos from open class websites, then extract frames from the videos, and then annotate the video frames, including object detection annotation and image classification annotation. Next, input the annotated data into the object detection model and the LVLM for fine-tuning training. After the training is completed, deploy the model to the project server. In the actual application process, collect the problems feedback from the market, and carry out iterative optimization on the video data, annotation data and training methods according to the feedback problems.

Existing student classroom behavior detection algorithms can be roughly divided into three classes: video-action-recognition-based\cite{huang2022multi}, pose-estimation-based\cite{he2020recognition} and object-detection-based\cite{yan2023student}. Video-based student classroom behavior detection enables the recognition of continuous behavior, which requires labeling a large number of samples. For example, the AVA dataset\cite{gu2018ava} for SlowFast\cite{feichtenhofer2019slowfast} detection is annotated with 1.58M. And, video behavior recognition detection is not yet mature, as in UCF101\cite{soomro2012ucf101} and Kinetics400\cite{carreira2017quo}, some actions can sometimes be determined by the context or scene alone. Pose-estimation-based algorithms characterize human behavior by obtaining position and motion information of each joint in the body, but they are not applicable for behavior detection in overcrowded classrooms. Considering the challenges at hand, object-detection-based algorithms present a promising solution. In fact, in recent years object-detection-based algorithms have made tremendous breakthroughs, such as YOLOv7\cite{wang2023yolov7}. Therefore, we have employed an object-detection-based algorithm in this paper to analyze student and teacher behavior. In addition, from the market feedback, this paper finds that some users do not pay attention to the behavior of each student in the current video frame, but care more about the group behavior of the overall students. Based on this, this paper adopts the method of image classification to classify the student behaviors and teacher behaviors in the video frames, and uses LVLM for fine-tuning training.

Based on real classroom scenarios, this paper proposes the Sudent and Teacher Classroom Behavior Dataset (SCB-Dataset), which contains a total of xx images and xx labels. It is divided into two types of datasets: object detection and image classification datasets. The two datasets have x and y behavior classes respectively, and both include student and teacher behaviors. The advantage of the object detection dataset is that it can locate the coordinates of each student and teacher and provide behavior classification information, which provides supporting data for subsequent fine-grained research. The advantage of the image classification dataset is that the workload is much smaller, and it can achieve behavior recognition that is difficult to complete by object detection. For example, the teacher's patrol and the student's listening depend on the overall information of the image, which is difficult to complete the classification through the local features of a single object.

\begin{figure*}[htbp]
\centerline{\includegraphics[width=1\textwidth]{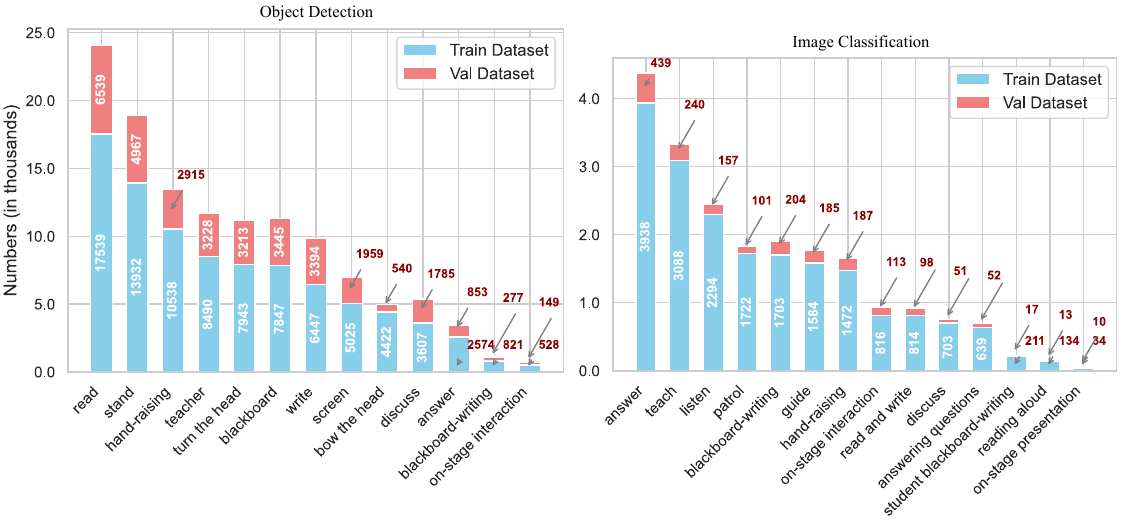}}
\caption{SCB-Dataset Class Count Statistics}
\label{Category-bar-chart}
\end{figure*}

The main contributions of this paper are as follows:

1) To the best of our knowledge, SCB-Dataset is the first public dataset of student and teacher classroom behaviors in real educational environments with the most classes and the largest number of images. The proposed dataset fills the gaps in student classroom behavior research under teaching scenarios.  

2) The dataset is divided into two types. The first is an object detection dataset, which can be used to accurately locate the positions of students and teachers and classify behaviors, suitable for traditional deep convolutional neural network models. The second is an image classification dataset, which is used to classify the behaviors of students and teachers in the current image as a whole, suitable for the latest LVLM.  

3) SCB-Dataset has been tested and analyzed in object detection series models and LVLM, providing baseline data references for follow-up research.

\begin{figure}[t]
\centerline{\includegraphics[width=0.5\textwidth]{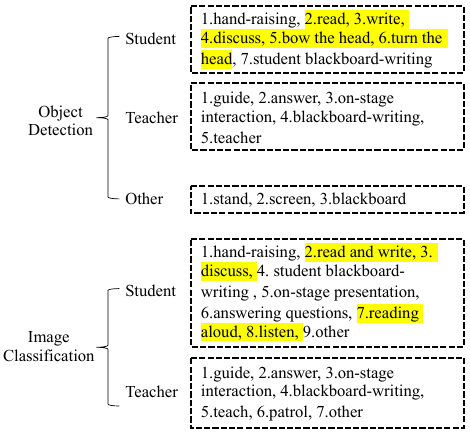}}
\caption{The behavior classification of SCB-Dataset}
\label{classes}
\end{figure}

\begin{figure}[t]
\centerline{\includegraphics[width=0.5\textwidth]{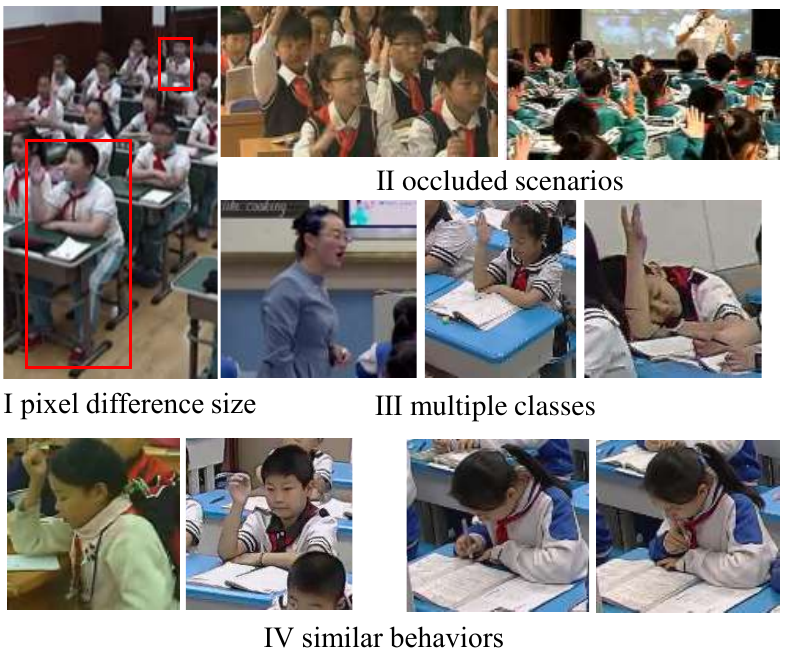}}
\caption{Challenges in the SCB-Dataset include pixel differences, dense environments, the coexistence of multiple classes , and similar behaviors.}
\label{SCB-Challenges}
\end{figure}

\begin{figure}[htbp]
\centerline{\includegraphics[width=0.5\textwidth]{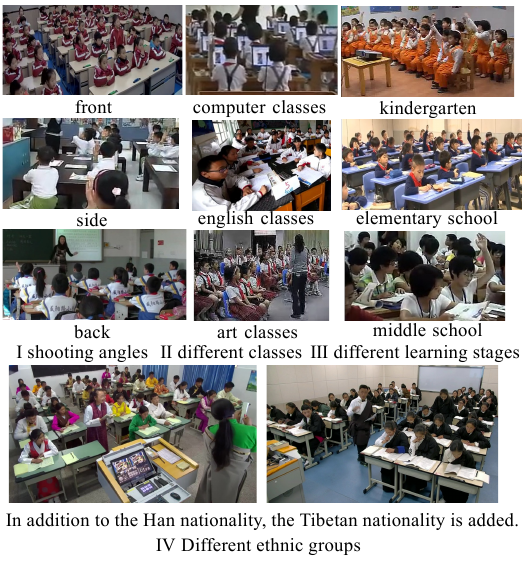}}
\caption{The diversity of the SCB-Dataset includes varying shooting angles, class differences, different learning stages, and different ethnic groups.}
\label{diversity_SCB-Dataset}
\end{figure}

\section{Related Work}
\subsection{Student Behavior Dataset}
In recent years, many researchers have adopted computer vision technology to automatically detect students' classroom behaviors. Meanwhile, a number of open-source and closed-source student behavior datasets have emerged, though the open-source ones only account for a small proportion.  As of now, this paper has collected 6 publicly available datasets, including: STBD-08 \citep{zhao2023cbph}, ClaBehavior \citep{wang2023students}, SCBehavior \citep{wang2024sbd}, UK\_Datasets \cite{feng2025imrmb}, the universe roboflow website \citep{classroom-dodzk_dataset}, and a student classroom behavior dataset from a paid website \citep{student_headup_detection2023}, this paper finds that the three datasets—STBD-08, UK\_Datasets, and the paid website dataset—are basically identical in terms of image content, sample quantity, and classification criteria. Furthermore, they contain numerous problematic data points, such as non-standard bounding boxes (bbox) and class labeling errors. Notably, both ClaBehavior and SCBehavior only provide 400 publicly available images, which is far fewer than the quantity stated in their respective papers.

The types of student classroom behavior datasets are mainly object detection data \citep{lu2025pacr}, with a small portion being human skeleton key points data \citep{zhou2023stuart}. There are also video action recognition data, video action captioning (description) data \citep{sun2021student}, and image action classification data. In terms of behavior classification, there are numerous classes \citep{yang20239student}, including common ones such as listening, hand-raising, reading, bending over/sleeping, writing, standing, using mobile phones, discussing, turning head. There are also some less common classes \citep{peng2025yolo}, such as looking up, guiding, focus, distract, playing, writing on the blackboard, teacher, etc. For more details on Open-source and Closed-source datasets, please refer to Appendix \ref{Open Source Dataset} and Appendix \ref{Closed-source Dataset}.

\subsection{Detection Algorithm}
Existing student behavior detection algorithms can be classified into three categories: video-action-recognition-based \citep{huang2022multi}, pose-estimation-based \citep{he2020recognition}, and object-detection-based \citep{yan2023student}. In recent years, due to significant advancements in the field, object-detection-based methods have emerged as the mainstream approach. Although video-based detection allows for the recognition of continuous behavior, it requires a large number of annotated samples such as in the AVA dataset \citep{gu2018ava} for SlowFast \citep{feichtenhofer2019slowfast} detection which includes 1.58M annotations. Pose-estimation algorithms obtain joint position and motion information but are not adequate for detecting behavior in overcrowded classrooms. Recently, the revolutionary changes brought about by Large vision-language models ( LVLMs )  worldwide \citep{bai2025qwen2} have also made them a popular area of research in student behavior detection.

\section{SCB-Dataset}
In this section, we describe SCB-Dataset, including its collection process, classification annotation information, statistical analysis, and existing challenges.

\subsection{Data Collection}

Aiming to ensure the dataset's diversity for real classroom scenarios, the dataset was directly collected from websites: bilibili, TikTok, 1s1k, and bjyhjy. Notably, classroom data from China's ethnic minorities were also incorporated. Subsequently, the collected videos were subjected to frame extraction. To reduce the imbalance among behavior classes, a differential frame selection strategy was adopted, reducing the sampling volume for common classes such as "read" and "write", and increasing the sampling volume for rare classes such as "discuss" and "board writing".


\subsection{Behavior classes}

The behavior classification of SCB-Dataset is shown in Fig.~\ref{classes}. The Object Detection dataset contains 12 types of behaviors in total, and the Image Classification dataset has 14 types of behaviors. By removing the overlapping behaviors between the two, SCB-Dataset has a total of 19 types of behaviors. In classroom settings, this paper holds that the analysis of student behaviors can be carried out from two perspectives: group and individual. The behavior classes marked with a yellow background in Fig.~\ref{classes} are the group behaviors of students.


Specifically, hand-raising is considered an individual behavior, where one or more students raise their hands to answer questions when the teacher asks, this behavior can reflect the teacher's questioning style, teacher-student interaction, and the level of classroom activity. Reading and writing are classed as group behaviors, where the reading and writing behaviors of an individual or a few students lack representativeness. It is more crucial to focus on the overall classroom engagement of students. The reading and writing behaviors of the majority can better mirror the current teaching dynamics in the classroom.



\begin{figure}[t]
\centerline{
\includegraphics[width=0.5\textwidth]{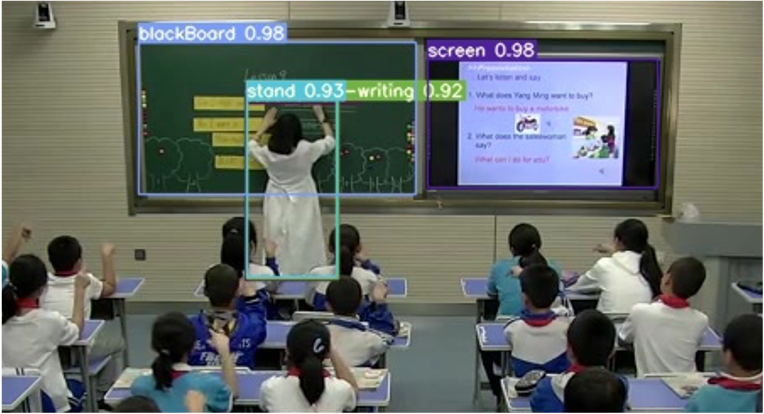}
}
\caption{Example of YOLOv7 detection results}
\label{YOLOv7_detection}
\end{figure}

In the Object Detection dataset, there are two types of "blackboard-writing": "student blackboard-writing" and "teacher blackboard-writing". These two behaviors have completely different meanings in classroom evaluation. "Student blackboard-writing" represents students' stage demonstration links, and "teacher blackboard-writing" is one of the important teaching links for teachers. In the teacher behavior class, "teacher" belongs to identity recognition. It is worth noting that the teacher behavior class in the Object Detection dataset does not have behaviors such as "teach" and "patrol". This paper tests and finds that the YOLO series network is difficult to fit their behavior characteristics because these behaviors need to be combined with environmental characteristics or timing characteristics. In other, "stand" can belong to the behavior of both teachers and students, so it is classified as Other. "screen" and "blackboard" are not behaviors and are classified into other. The role of these two classes is to obtain the teacher's blackboard-writing content in the classroom scene and the content such as PPT in the screen.

\begin{table}[htbp]
\label{Intervention-string}
\centering
\renewcommand{\arraystretch}{1.4}
\begin{tabular}{lcccc}

    \toprule
     \multirow{2}{*}{Dataset} & \multicolumn{2}{c}{Object Detection} & \multicolumn{2}{c}{Image Classification} \\
    \cline{2-5}  
    & Images & Annotation & Images & Annotation \\
    \midrule
    Train & - & 89,713 & 19,152  & 19,152 
    \\ 
    Val & - & 33,264 & 1,867 & 1,867 \\
    Total & 13,330 &  122,977 & 21,019 & 21,019\\

\bottomrule
\end{tabular}
\caption{Statistics on the number of images and annotations of the two datasets in SCB-Dataset}
\end{table}

In the Image Classification dataset, since this paper uses LVLMs for fine-tuning training, two sets of prompts for students and teachers are used. For this reason, both sets of prompts contain "other". To enable LVLMs to better learn behavioral characteristics, in the image screening of the "hand-raising" category, only images containing 3 or more hand-raising behaviors are retained. In addition, because the image classification in this paper outputs only one class per image, adjustments have been made to the classes, with the "read" and "write" behaviors merged. The greatest advantage of Image Classification is that it can combine the information of the entire image to identify some behaviors that are difficult to define or recognize in object detection, such as: "on-stage presentation", "reading aloud", "listen", "teach", "patrol".

\begin{table}[t]
\fontsize{9pt}{11pt}\selectfont
\centering

\renewcommand{\arraystretch}{1.4}
\begin{tabular}{lcccc}

    \toprule
    \textbf{class} & P & R & mAP@0.5 & mAP@.95 \\
    \midrule
    hand-raising & 79.4 & 76.9 & 79.2 & 59.4  \\
    read & 65.5 & 68.2 & 70.5 & 52.9 \\
    write & 68.4 & 67.8 & 72.2 & 58.1 \\
    discuss & 67.5 & 72.5 & 74.7 & 39.3 \\
    bow the head & 26.0 & 34.0 & 21.9 & 7.9 \\
    turn the head & 23.8 & 44.0 & 23.5 & 9.5 \\
    guide & 88.5 & 78.3 & 83.6 & 48.9 \\
    answer & 86.2 & 86.6 & 91.5 & 80.8 \\
    on-stage interaction & 82.3 & 84.5 & 90.1 & 81.5 \\
    blackboard-writing & 91.0 & 93.5 & 96.4 & 86.6 \\
    teacher & 95.5 & 95.2 & 97.7 & 83.0 \\
    stand & 93.1 & 94.7 & 96.6 & 79.8 \\
    screen & 96.1 & 97.1 & 97.9 & 92.5 \\
    blackboard & 96.2 & 97.1 & 98.1 & 93.3 \\
\bottomrule
\end{tabular}
\caption{Training results of YOLOv7 on the object detection dataset in SCB-Dataset}
\label{Object-Detection-SCB-Dataset-YOLOv7}
\end{table}

\subsection{Dataset challenges}


Classrooms are densely populated environments, which also bring many challenges to SCB-Dataset. For example, as shown in Fig.~\ref{SCB-Challenges} \uppercase\expandafter{\romannumeral 1}, there is a significant pixel difference between the images of students in the front row and those in the back row. As shown in Fig.~\ref{SCB-Challenges} \uppercase\expandafter{\romannumeral 2}, the dense students lead to serious front and rear occluded situations. As shown in Fig.~\ref{SCB-Challenges} \uppercase\expandafter{\romannumeral 3}, students/teachers may have multiple behaviors at the same time: teachers who stand and teach, students who are hand-raising and reading, which is called "multiple classes". As shown in Fig.~\ref{SCB-Challenges} \uppercase\expandafter{\romannumeral 4}, there is a high degree of similarity between behaviors, such as the similarity between placing a hand on the forehead and raising a hand, or the similarity between writing and reading.

The SCB-Dataset exhibits a rich diversity, as shown in Fig.~\ref{diversity_SCB-Dataset} \uppercase\expandafter{\romannumeral 1}, encompassing a variety of perspectives within classroom settings, including frontal, lateral, and back views. The same behavior can significantly differ when viewed from various angles, which increases the complexity of behavior detection. As demonstrated in Fig.~\ref{diversity_SCB-Dataset} \uppercase\expandafter{\romannumeral 2}, the dataset also includes a range of classroom environments and course types, for instance, computer courses are typically conducted in well-equipped computer labs, while English and other cultural courses are held in standard classrooms, and art courses might take place in orderly arranged rehearsal rooms. As presented in Fig.~\ref{diversity_SCB-Dataset} \uppercase\expandafter{\romannumeral 3}, the dataset covers students' growth stages from kindergarten through university, and as shown in Fig.~\ref{diversity_SCB-Dataset} \uppercase\expandafter{\romannumeral 4}, it includes the diversity of different ethnic backgrounds. This comprehensive consideration across ages, cultures, and environments provides a more thorough and in-depth data foundation for research.

\begin{figure}[t]
\centerline{
\includegraphics[width=0.5\textwidth]{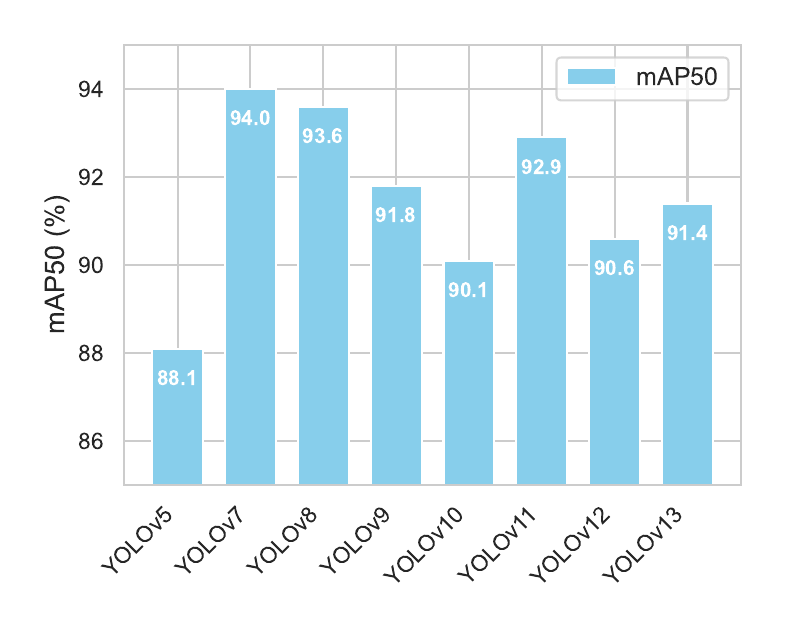}
}
\caption{
Training and testing results of the SCB-Dataset dataset (teacher behavior part) on YOLO series models}
\label{yolov_mAP50_comparison}
\end{figure}

\subsection{Dataset statistics}


As shown in Fig.~\ref{Category-bar-chart}, this paper counts the number of class annotations of two datasets. The bar chart on the left of the figure is the class statistics of the Object Detection dataset, and the one on the right is the class statistics of the Image Classification dataset. It can be seen from the figure that although efforts have been made to alleviate the problem of class data imbalance during data collection, the problem is still serious according to the statistical results. In addition, it can be seen that the number of annotations of the Object Detection dataset is much higher than that of the Image Classification dataset. This is because the former has multiple objects in a single image, and each object has multiple classification annotations, while the latter has only one classification for a single image. 

This paper also counts the total number of annotations and the number of images of the two datasets, as shown in Table 2. It should be noted that the object detection dataset does not separately record the number of training sets (train) and validation sets (val) for the following reasons: There is a serious class imbalance problem in this dataset. For example, the number of samples of the "read" and "write" classes is much larger than that of the "discuss" class. If all the objects of the "read" and "write" classes in all images are annotated, the imbalance will be further aggravated. Therefore, SCB-Dataset only annotates "read" and "write" in some images, while all annotations are made for the "discuss" class. This processing makes SCB-Dataset split into multiple sub-parts, and the training sets and validation sets of each sub-part are independently and randomly divided in a ratio of 4:1, resulting in overlaps between the training (train) sets and validation (val) sets of different sub-parts. Based on the above situation, the overall number of training sets and validation sets of the object detection dataset has no practical reference significance, and only the separate analysis of the division data of each sub-part has statistical value (see Appendix x for details).

\section{Experiment}
This experiment mainly conducts benchmark tests on SCB-Dataset with Object Detection models and LVLM. 


\begin{figure}[t]
\centerline{
\includegraphics[width=0.5\textwidth]{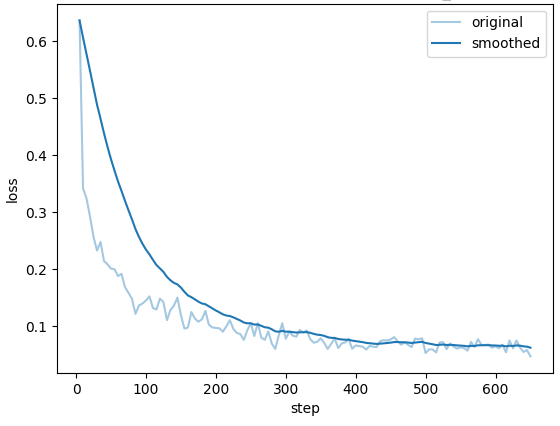}
}
\caption{Loss of Qwen2.5-VL-7B-instruct during the training iteration process}
\label{LVLM_training_loss}
\end{figure}


\subsection{Experimental setup}

\textbf{Environment}
This experiment was conducted using an NVIDIA vGPU-32GB GPU with 32GB of video memory and a CPU12 vCPU Intel(R) Xeon(R) Platinum 8352V CPU, with Ubuntu 22.04 as the operating system. The software versions included PyTorch 2.3.0, Python 3.12, and CUDA 12.1.

\textbf{Dataset}
The dataset used in our experiments is SCB-Dataset including Object Detection Dataset and Image Classification Dataset.


\textbf{Model Training}

The Object Detection Dataset in SCB-Dataset is trained using the YOLO series, with the epoch ranging from 30 to 120, the batch size set to 4, and the image size set to 640x640. The Image Classification Dataset uses the LLaMA Factory framework to train LVLM, the LVLM uses Qwen2.5-VL-7B-instruct, adopts the LoRA method, learning rate set to $5\times10^{-5}$, number of training epochs set to 2, 
batch size set to 2, LoRA rank set to 8, scaling factor set to 16, dropout rate set to 0.1, and LoRA+ learning rate ratio set to 16.




\begin{table}[t]
\fontsize{9pt}{11pt}\selectfont
\centering
\renewcommand{\arraystretch}{1.4}

\begin{tabular}{lccc}

    \toprule
    \textbf{class} & P & R & f1 \\
    \midrule
    hand-raising & 87.0 & 85.6 & 86.3  \\
    read and write & 83.6 & 93.9 & 88.5 \\
    discuss & 93.9 & 90.2 & 92.0  \\
    student blackboard-writing & 83.3 & 88.2 & 85.7  \\
    on-stage presentation & 100 & 70.0 & 82.4 \\
    answering questions & 75.0 & 69.2 & 72.0  \\
    reading aloud & 100 & 69.2 & 81.8  \\
    listen & 88.1 & 89.2 & 88.6  \\
    guide & 87.0 & 50.8 & 64.2  \\
    answer & 87.6 & 83.4 & 85.4  \\
    on-stage interaction & 89.2 & 73.5 & 80.6  \\
    blackboard-writing & 99.0 & 98.5 & 98.8 \\
    teach & 87.4 & 92.1 & 89.7  \\
    patrol & 42.3 & 87.1 & 57.0  \\
    \midrule
    all & 86.1 & 83.4 & 83.8 \\
\bottomrule
\end{tabular}
\caption{Training results of Qwen2.5-VL-7B-instruct on the image classification dataset in SCB-Dataset}
\label{Image_Classification_SCB_Dataset_LVLMs}
\end{table}

\subsection{Object Detection}
This experiment uses YOLOv7 to conduct benchmark tests on the Object Detection dataset of SCB-Dataset. In addition, it also uses the teacher behavior part of the dataset to test the baselines of v5, v8, v9, v10, v11, v12, v13, and finally finds that YOLOv7 has the best effect.

Table \ref{Object-Detection-SCB-Dataset-YOLOv7} shows the training results of YOLOv7 on the object detection dataset in SCB-Dataset, with the data units in the table being "\%", precision represented by "p", recall represented by "R", mAP@0.5 representing mean Average Precision at Intersection over Union threshold of 0.5, and mAP@0.95 representing mean Average Precision at Intersection over Union threshold of 0.95. As can be seen from Table \ref{Object-Detection-SCB-Dataset-YOLOv7}, except for bow the head and turn the head, the mAP@0.5 of other behaviors are all above 70\%, among which even half of the behaviors have mAP@0.5 reaching above 90\%. This paper holds that as long as mAP@0.5 reaches 70\%, the behavior can basically be used for practical applications. Examples of YOLOv7 detection results can be seen in Fig.~\ref{YOLOv7_detection}.




This experiment also conducted training and validation on the SCB-Dataset dataset (teacher behavior part) on YOLO series models. As shown in Fig.~\ref{yolov_mAP50_comparison}, it can be found that the lowest mAP50 is YOLOv5, which is only 88.1\%, the highest is YOLOv7, reaching 94\%, followed by YOLOv8 with 93.6\%, but the performance of YOLOv9$\sim$v13 launched in recent years is generally average.


\subsection{Image Classification}
\begin{figure}[t]
\centerline{
\includegraphics[width=0.5\textwidth]{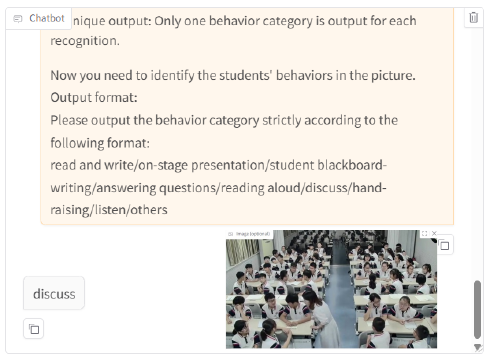}
}
\caption{Using the LLaMA Factory framework to test the Qwen2.5-VL-7B-instruct example}
\label{Qwen2.5-VL-7B_sample}
\end{figure}

This experiment uses the LLaMA Factory framework to train Qwen2.5-VL-7B-instruct, Table \ref{Image_Classification_SCB_Dataset_LVLMs} shows the training results of Qwen2.5-VL-7B-instruct on the image classificat dataset in SCB-Dataset. It can be seen that the f1 of behaviors are almost all above 80\%, and the training effect is significant. Fig.~\ref{LVLM_training_loss} shows the decreasing process of loss during training. There are 650 iterations in total. The loss decreases rapidly in the first 150 iterations, tends to be stable in the last 200 iterations, and finally decreases to 0.0471. Fig.\ref{Qwen2.5-VL-7B_sample} shows an example of testing Qwen2.5-VL-7B-instruct using the LLaMA Factory framework (Using the LLaMA Factory framework to test the Qwen2.5-VL-7B-instruct example). When the user inputs an image and a prompt, Qwen2.5-VL-7B-instruct will provide the corresponding classification for the image.

\section{Conclusion}


In summary, this paper fills the gap in student-teacher behavior datasets in this field through the construction of SCB-dataset and its evaluation using YOLO series algorithms and LVLM, and provides benchmark test results. SCB-dataset is helpful to promote the application of artificial intelligence in education, improve teaching efficiency, etc. We are also continuously expanding the scale of the dataset to adapt to various challenges in real environments.

\bibliography{aaai2026}

\appendix


\section{Appendix}

\subsection{Annotation Work}
Annotation work is the most time-consuming and labor-intensive part of SCB-Dataset, accounting for nearly 90\% of the total workload. Since 2021, we have gone through the entire process, from defining input-output expectations and behavior classifications to formulating annotation rules. However, due to numerous unreasonable and immature definitions in the early stages of dataset creation, almost a year and a half of time was wasted. By the first half of 2023, we redesigned the annotation process and introduced an extensible behavior annotation method, which allows us to flexibly expand on the existing foundation, no matter how many behaviors need to be added in the future.

\textbf{Improvements to Annotation Tools}

To meet the practical needs of annotation work, we made multiple versions of optimizations and improvements to the annotation tool VIA. The details are as follows:

\textbf{VIA Original Version}

The link to the original version of VIA is as follows(as shown in Fig.~\ref{VIA0} ):
\url{https://whiffe.github.io/VIA/via_image_annotator.html}

\begin{figure}[htbp]
\centerline{\includegraphics[width=0.5\textwidth]{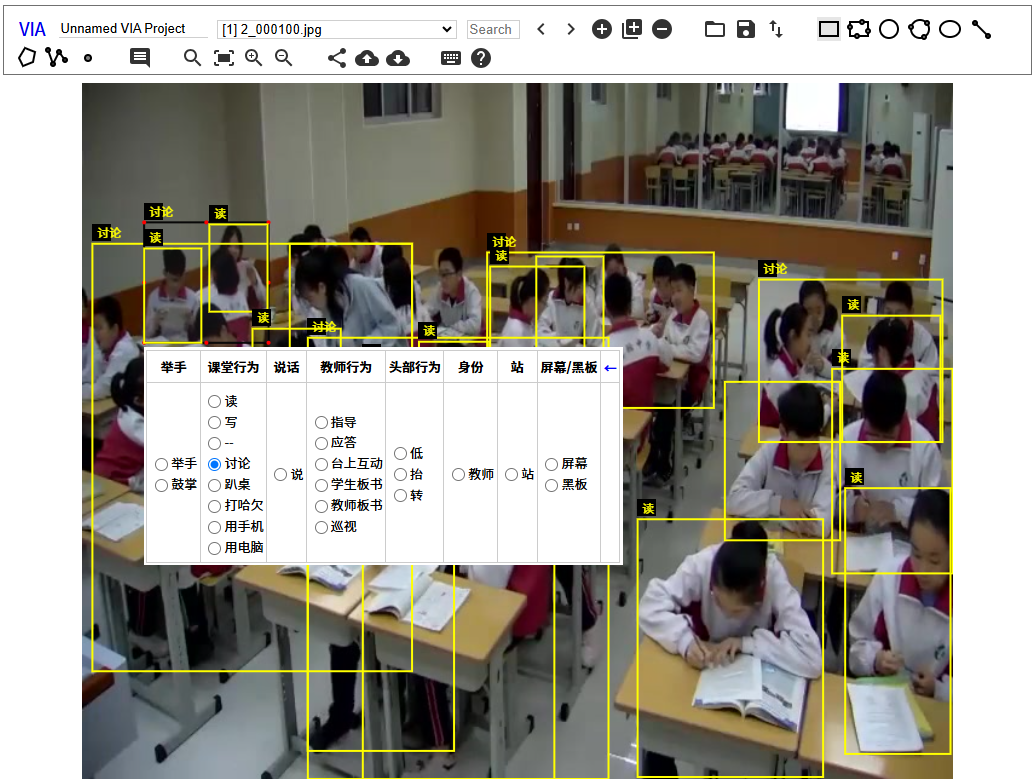}}
\caption{Original Annotation Interface of VIA}
\label{VIA0}
\end{figure}

\textbf{Second Version}

\begin{figure}[htbp]
\centerline{\includegraphics[width=0.5\textwidth]{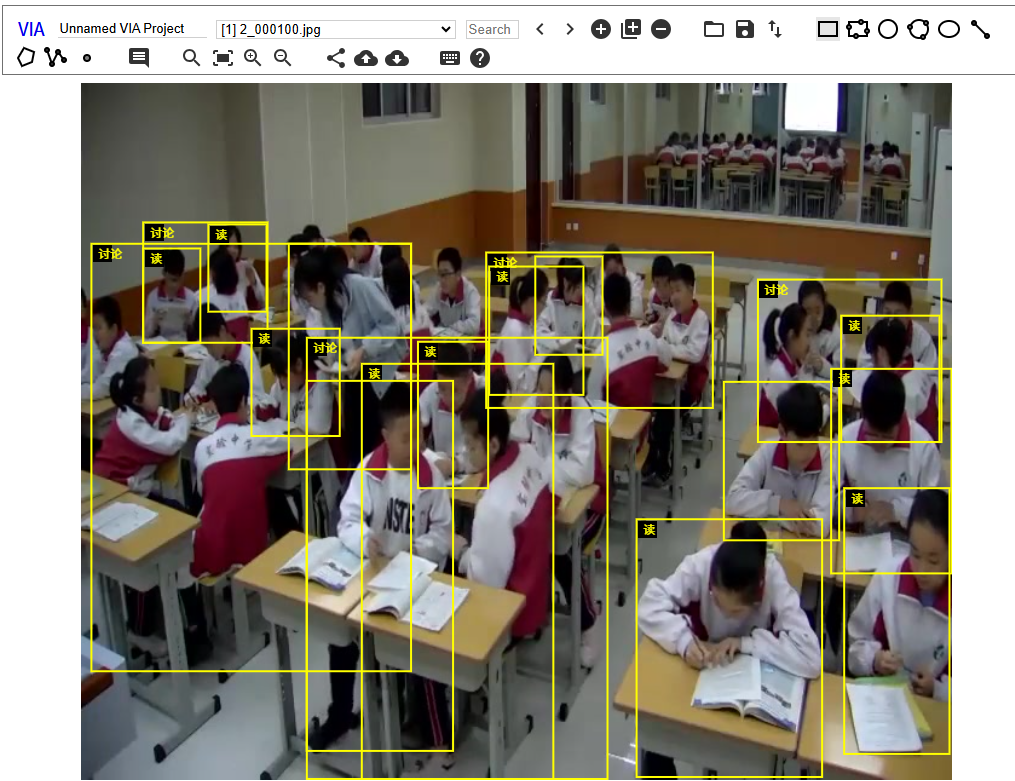}}
\caption{VIA Second Version}
\label{VIA1}
\end{figure}

In this version, we optimized the label display position. As shown in Fig.~\ref{VIA1}. Labels are now displayed inside the annotation boxes instead of outside. This improvement was designed for classroom scenarios where many annotation boxes are located at the top of the image. Displaying labels inside the boxes makes it more convenient for inspection and verification. We have further optimized the function of switching the display of labels. In addition to using the mouse scroll wheel, we have added the keys "z" and "x" to switch the display of labels. This means that users can switch the display of labels either by using the mouse scroll wheel or by pressing the keys "z" and "x". This design allows users to easily switch labels even without a mouse, making the use more convenient.
\url{https://whiffe.github.io/VIA/via_image_annotatorK.html}

\textbf{Third Version}

\begin{figure}[htbp]
\centerline{\includegraphics[width=0.5\textwidth]{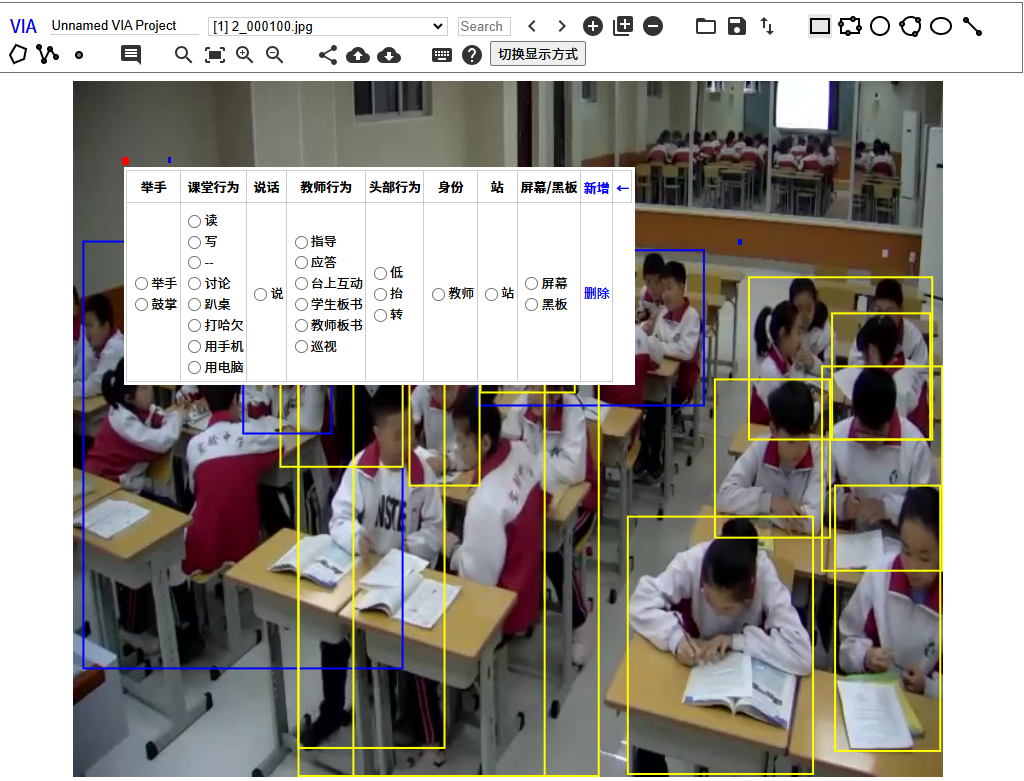}}
\caption{VIA Third Version}
\label{VIA2}
\end{figure}

The third version further optimized the selection of annotation boxes by introducing the mouse scroll switching feature. As shown in Fig.~\ref{VIA2}. In the original version of VIA, annotation boxes could only be selected by clicking with the mouse. If the annotation box was too small (typically caused by mislabeling), it became difficult to select. This version is particularly suitable for cleaning up small boxes created by mislabeling. Additionally, when scrolling the mouse, the selected annotation box changes color, helping users identify which boxes have been selected and which have not. This feature is especially useful for images containing a large number of targets.
\url{https://whiffe.github.io/VIA/via_image_annotator2.html}

\textbf{Fourth Version}  

\begin{figure}[htbp]
\centerline{\includegraphics[width=0.5\textwidth]{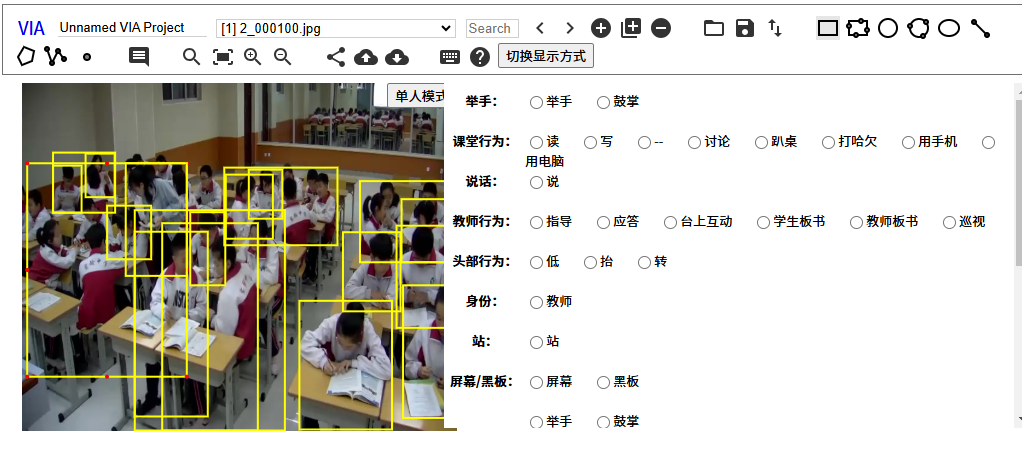}}
\centerline{\includegraphics[width=0.5\textwidth]{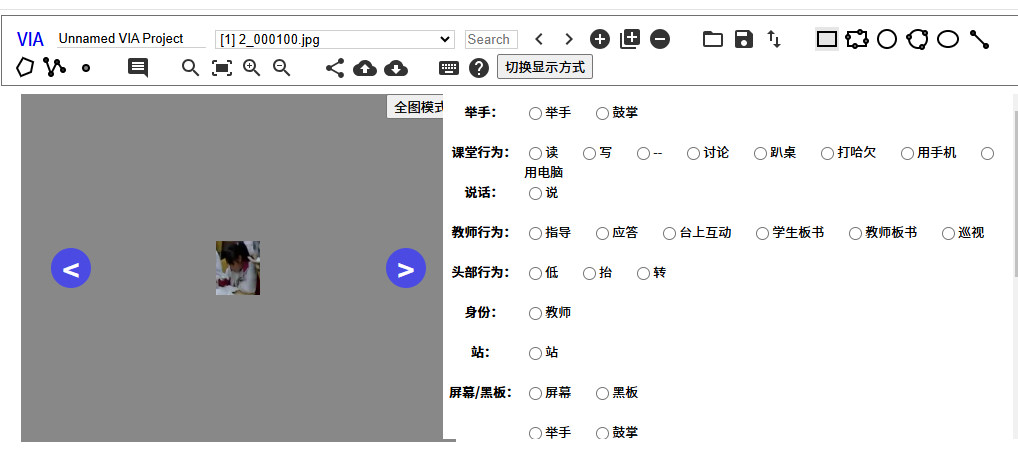}}
\caption{VIA Fourth Version}
\label{VIA34}
\end{figure}

In the fourth version, as shown in Fig.~\ref{VIA34}. We optimized the display of annotation content by showing it in half-page format, which significantly improves annotation efficiency. Additionally, we introduced two new modes: \textbf{Full Image Mode} and \textbf{Single Target Mode}.  
\begin{itemize}
    \item \textbf{Full Image Mode}: Displays all annotation boxes in the entire image. 
    \item \textbf{Single Target Mode}: Displays each annotated target individually. This feature is particularly suitable for dense scenarios, allowing users to check whether each annotation box is accurately drawn and aiding in behavior classification verification and analysis.  
\end{itemize}
\url{https://whiffe.github.io/VIA/via_image_annotator3.html}

\textbf{Fifth Version}

\begin{figure}[htbp]
\centerline{\includegraphics[width=0.5\textwidth]{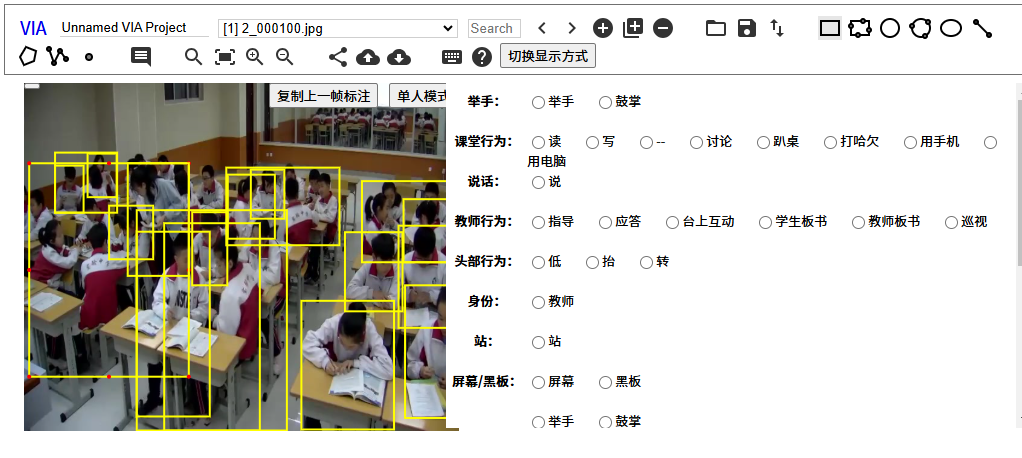}}
\caption{VIA Fifth Version}
\label{VIA5}
\end{figure}

Building on the previous version, the fifth version introduced the copy previous frame annotations feature. As shown in Fig.~\ref{VIA5}. This functionality is particularly useful for annotating consecutive frames with high similarity, significantly reducing repetitive operations, improving annotation efficiency, and further lowering labor costs.
\url{https://whiffe.github.io/VIA/via_image_annotator4.html}

\textbf{Final Checks with viaJson}

\begin{figure}[htbp]
\centerline{\includegraphics[width=0.5\textwidth]{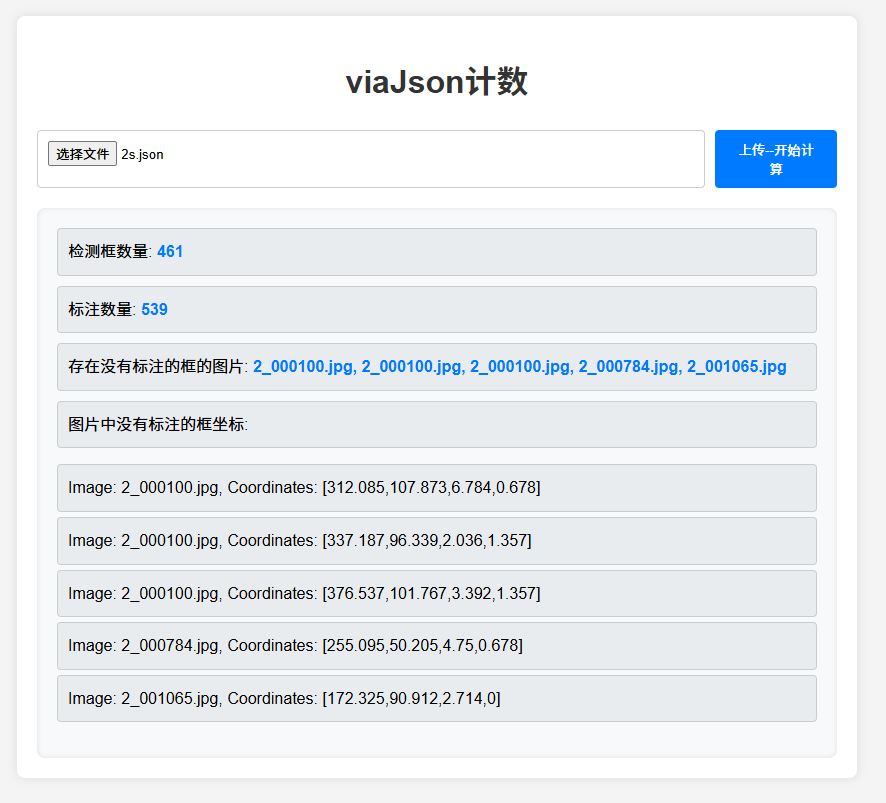}}
\caption{
Annotation Review and Counting Website}
\label{VIA6}
\end{figure}

After completing each annotation, we use the viaJson counting website to verify the annotation results. As shown in Fig.~\ref{VIA6}. This tool identifies any unclassified annotation boxes (i.e., boxes drawn but not categorized) and provides the coordinates of the annotation boxes. Additionally, it provides statistics on the number of detection boxes and annotated targets in the current file, helping us further ensure the completeness and accuracy of the annotations.
\url{https://whiffe.github.io/VIA/via_cout_labels.html}

\begin{figure*}[htbp]

\centerline{
\includegraphics[width=0.5\textwidth]{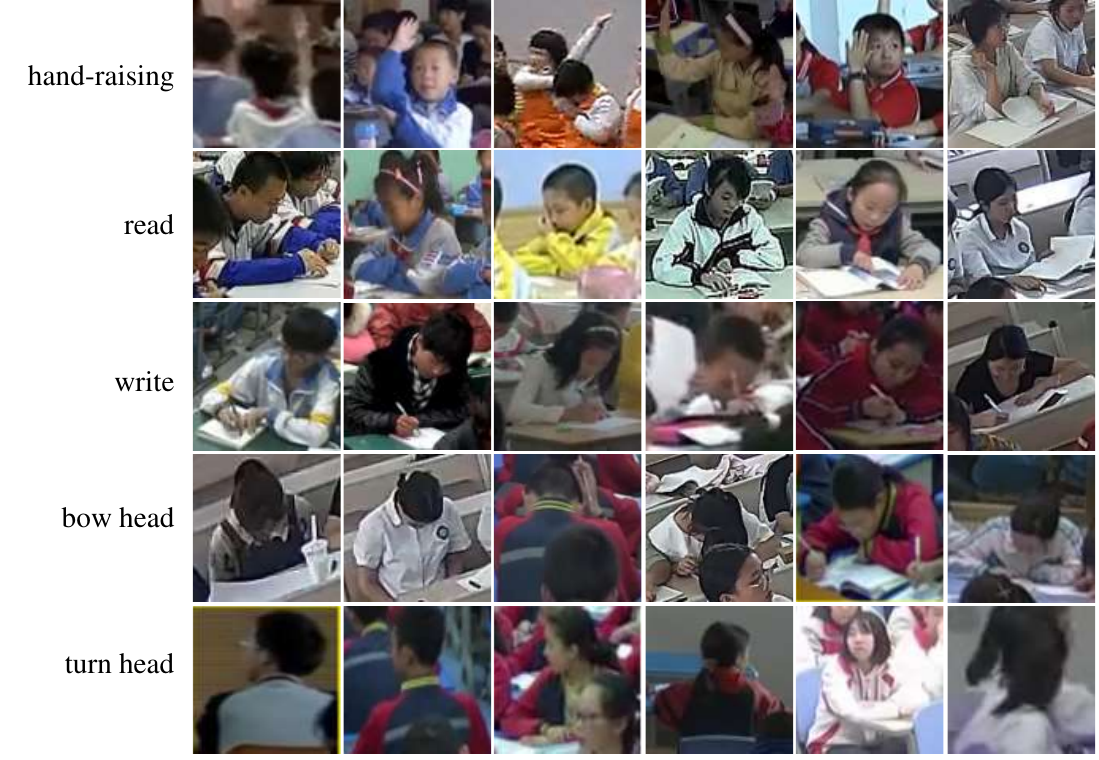}
\includegraphics[width=0.5\textwidth]{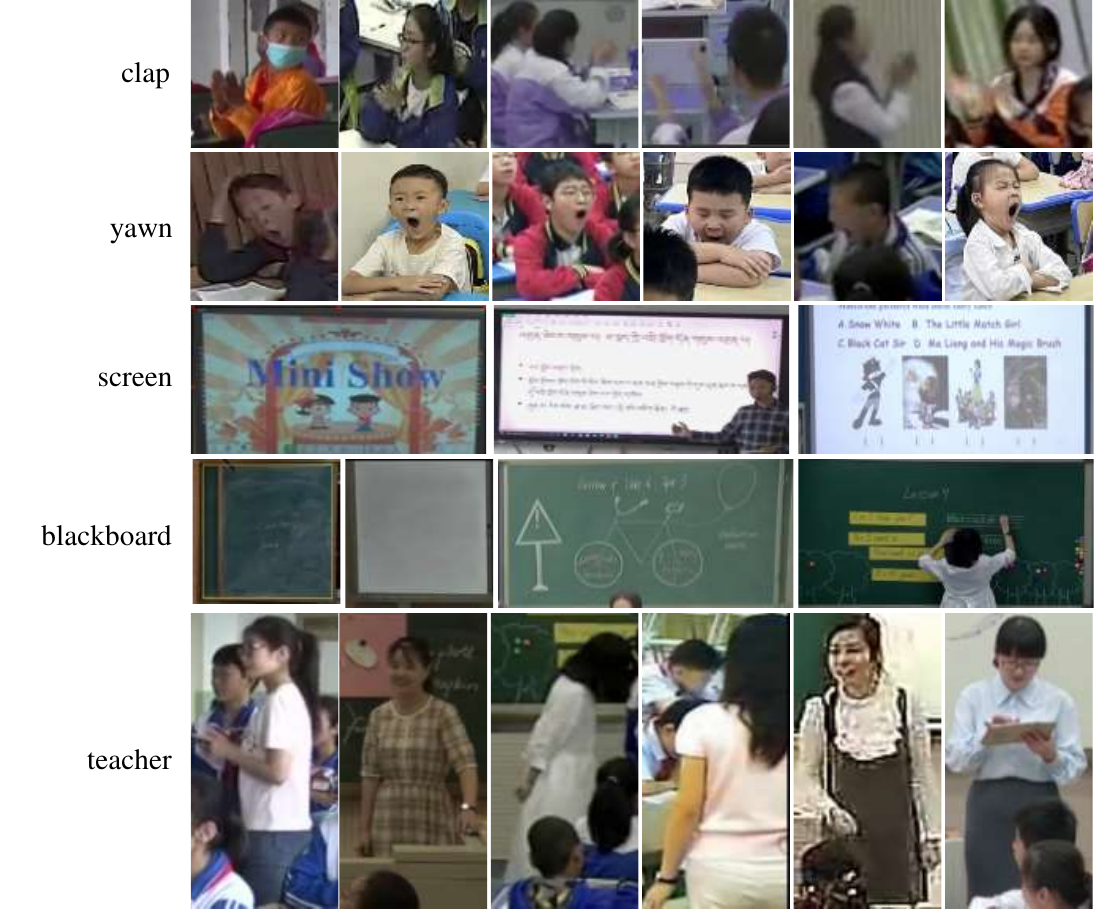}
}
\centerline{
\includegraphics[width=0.5\textwidth]{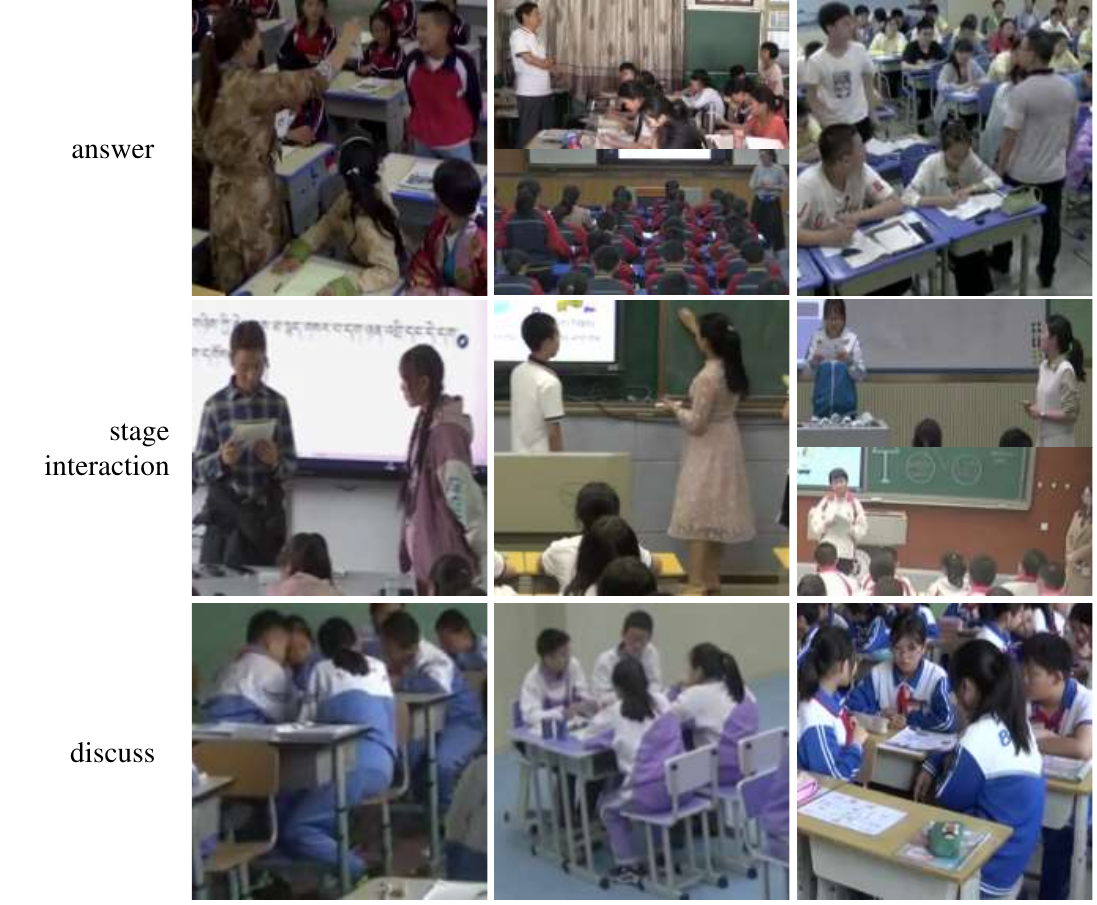}
\includegraphics[width=0.5\textwidth]{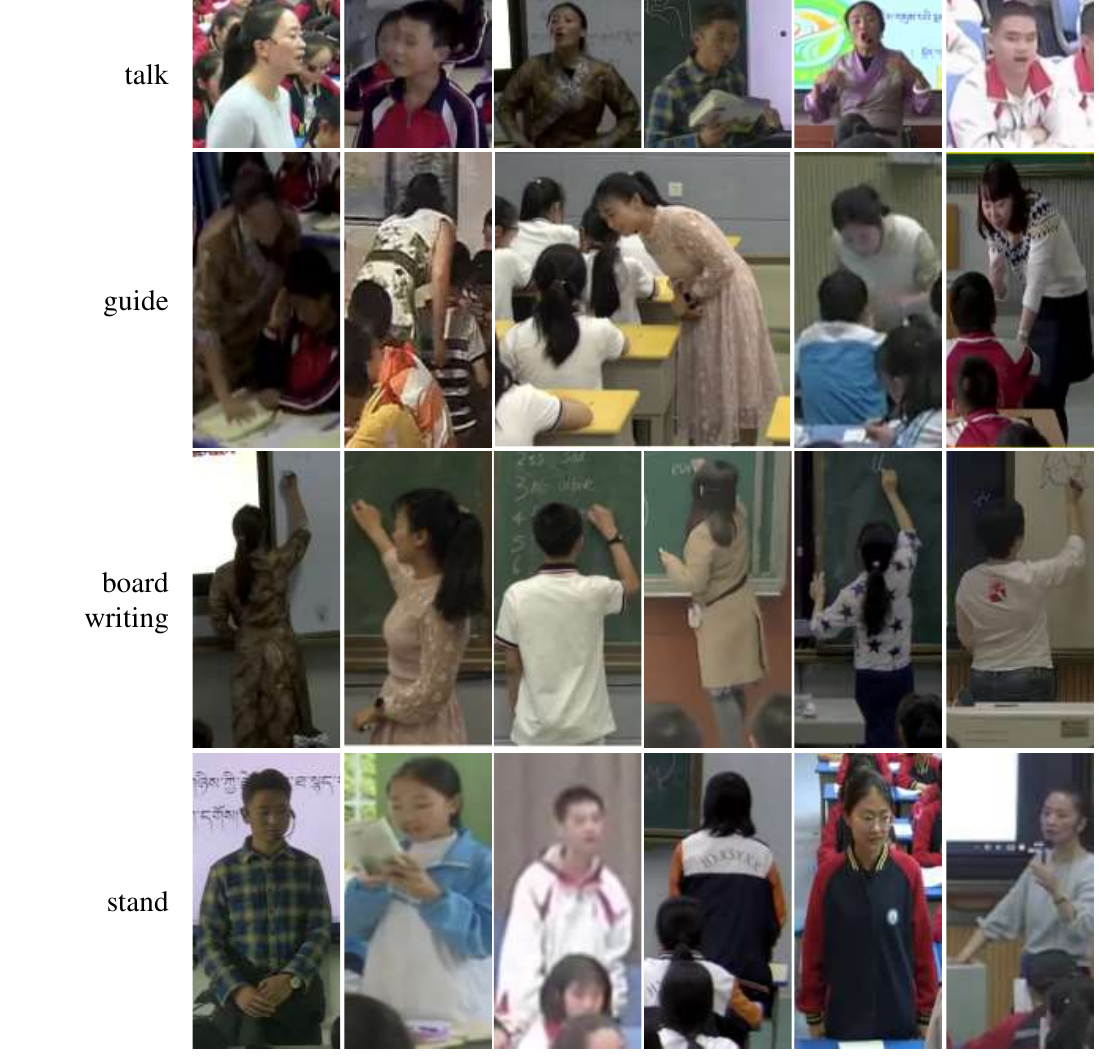}
}
\centerline{
\includegraphics[width=0.5\textwidth]{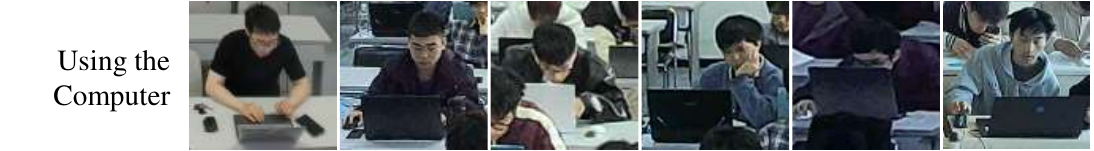}
\includegraphics[width=0.5\textwidth]{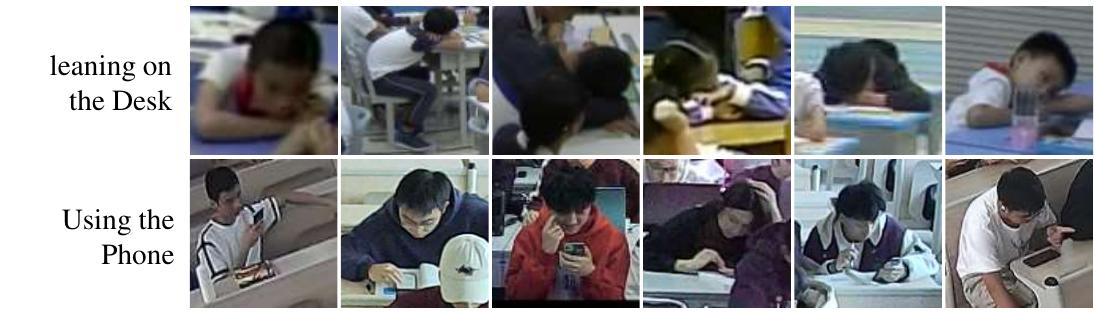}
}
\caption{Examples of images from various classes in the SCB dataset.}
\label{SCB5-1}
\end{figure*}

\subsection{Open Source and Closed Source Dataset}

\subsubsection{Open Source Dataset}
\label{Open Source Dataset}

\textbf{ClaBehavior}

The ClaBehavior paper\citep{wang2023students} mentions 1342 images and 9911 annotations. However, we actually found only 400 images and 8083 annotations on \url{run:https://github.com/CCNUZFW/Student-behavior-detection-system/tree/master/dataset/coco}{GitHub}, including:  Train Dataset: 360 images and 7,250 annotations, Val Dataset: 40 images and 833 annotations. The categories include: Write, Read, Lookup, Turn\_head, Raise\_hand, Stand, Discuss.

\begin{table}[htbp]
\centering  
\caption{ClaBehavior}
\label{ClaBehavior}
\begin{tabular}{llll}  
   \toprule
    & Train & Val & Total \\
   \midrule
   Write & 520 & 59 & 579 \\
   read & 920 & 101 & 1021 \\
   lookup & 4045 & 435 & 4480 \\
   turn\_head & 915 &  96 & 1011 \\
   raise\_hand & 569 & 115  & 684\\
   stand & 58 & 8  & 66\\
   discuss & 223 & 19  & 242\\
   \midrule
   Total & 7250 & 833 & 8083 \\
   \bottomrule
\end{tabular}
\end{table}

\textbf{STBD-08}
\label{stbd-08}

The STBD-08\citep{zhao2023cbph} paper states that the dataset contains 4432 images and 151574 annotations. However, when we conducted statistics on the data provided by the author, we found that the actual quantities far exceed these figures. Our statistics show that the dataset has 8884 images (including 7052 in the training set and 1,832 in the validation set) and 267888 annotations (including 212728 in the training set and 55160 in the validation set).

Through online search, we found that the STBD-08 dataset is completed based on the dataset publicly \url{https://mbd.pub/o/bread/ZZiTl5lw}{sale online} (the dataset also has 8,884 pieces), and the data volume is far lower than that of the dataset publicly sold online.

However, when we cleaned the STBD-08, we found that there were still many problem data in the STBD-08, such as the non-standard bbox, and some class labeling errors.

In other words, the author of STBD-08 has not made the dataset they created public, and only the original dataset purchased online is disclosed.

\begin{table}[htbp]
\centering  
\caption{STBD-08}
\label{STBD-08}
\begin{tabular}{llll}  
   \toprule
    & Train & Val & Total   \\
   \midrule
   Writing & 57164 & 15298 & 72462  \\
   Reading & 46872 & 12060 & 58932 \\
   Listening & 93509 & 24019 &  117528\\
   Turning around & 4314  & 1025 & 5339 \\
   Raising hand & 3336 & 847 & 4183 \\
   Standing & 3287 & 814 & 4101\\
   Discussing & 3710 & 953 & 4663\\
   Guiding & 536 & 144 & 680 \\
   \midrule
   Total & 212728 & 55160 & 267888  \\
   \bottomrule
\end{tabular}
\end{table}

\textbf{SCBehavior}

The SCBehavior paper\cite{wang2024sbd} mentions that there are 1346 images. However, when we checked the author's \url{https://github.com/CCNUZFW/SCBehavior}{GitHub}, we found only 400 damaged images that cannot be viewed (360 in the Train Dataset and 40 in the Val Dataset).

\textbf{UK\_Datasets}

UK\_Datasets\cite{feng2025imrmb} is derived from the 2019 elementary school classroom videos collected from the National Education Resources Public Service Platform (NERPSP). 

UK\_Datasets extracted 8754 images by frame, and considering the detection needs in real classroom scenarios, it classified these images into eight categories of typical student behaviors: writing, reading, listening, raising hands, turning, standing, discussing, and accepting teacher instructions.

The author categorized the test set portions of UK\_Datasets according to the degree of occlusion: "Heavy Occlusion (HO)"and "Low Occlusion (LO)".

Unfortunately, when we downloaded the \url{https://figshare.com/articles/dataset/IMRMB-Net_zip/27894246}(UK\_Datasets) for statistics, we found that the data was not original. Specifically, it originated from the \ref{stbd-08} section and the dataset publicly  \url{https://mbd.pub/o/bread/ZZiTl5lw}{sale online}  as introduced in this paper. The author merely divided and counted these existing datasets.

Since the data itself is plagiarized and not original, this paper will not conduct statistical analysis on its data.

\subsubsection{Closed-source Dataset}
\label{Closed-source Dataset}

There are many Closed-source datasets, as shown in Table~\ref{Close Source Dataset}.

\begin{table*}[htbp]
\centering
\caption{Close Source Dataset, A: Object Detection B: Human skeleton key points C: video action recognition D: Caption}
\label{Close Source Dataset}
\begin{tabular}{>{\raggedright\arraybackslash}p{4.5cm}|p{12cm}} 
    \toprule
    Dataset & Class and Static \\
    \midrule
    \cite{peng2025yolo} (A) & focus, distract, 1000 images \\
    \midrule
    CB Dataset\citep{dang2025object} (A)& listening (11,934),noting (8,727), playing (5,649), and groveling (2,977) \\
    \midrule
    HRSW Dataset \citep{lu2025pacr} (A) & rise hand, read, sleep, and write, 4,881 images, 1,2631 annotations.\\
    \midrule
    TCBDS \citep{ma2024improving} (A) & Teacher Classroom Behavior Data Set (TCBDS), facing the board (1,410), facing the students (1,415), writing on the board (1,034), teaching while facing the board(869), teaching while facing the students (978), and interactive (1,525). 6660 images ( 5,328 train images and 1,332 val images) \\
    \midrule
    SCB-E \citep{jiang2024scb} (A) & raising hands, reading, sleeping, writing, and using a mobile phone, 6,489 trainval images and 722 testing images  \\
    \midrule
    RSCB-Dataset \citep{jiang2024ldsbc} (A)& raising hands, reading, writing, sleeping, and using mobile phones, 5,221 images and 19,000 instances of specific behaviors.\\
    \midrule
    SB Dataset \citep{dang2024multi} (A)& listening (9,343), noting (7,243), playing (5,215), and grovelling (3,504) \\
    \midrule
    ActRec-Classroom \citep{fu2019learning} (AB)& listening carefully, hand raising to answer questions, participating in discussions, reading and note taking, 5126 images\\
    \midrule
    A large-scale dataset for student behavior \citep{zheng2020intelligent} (A)& hand-raising (70,000), standing (20,000), sleeping (3,000), 29,000 training images, 11,000 validate images \\
    \midrule
    BNU-LCSAD \citep{sun2021student} (ACD)& listening carefully (984), taking notes (582), using mobile phones (545), yawning (520), eating or drinking (515), reading (365), discussing (265), looking around (252), using computers (168), sleeping or snoring (80), and raising hands (15) \\
    \midrule
    Student Classroom Behavior Dataset \citep{zhou2022classroom} (B) & raising hands (10,000), bending over (10,00), walking back and forth(10,000), writing on the blackboard (10,000), looking up (10000), bowing their heads (10,000), standing (10,00), lying on their desks (1,000). \\
    \midrule
    Student behavior dataset \citep{li2023student} (A)& look at phone, listen to, stand, sleep, sit, talk, and write, 20,409 frames\\
    \midrule
    Student action dataset \citep{trabelsi2023real} & high and low attention, high: focused and raising hands, low: feeling bored, eating/drinking, laughing, reading, using a phone, distracted, and writing, 3,881 images \\
    \midrule
    A large-scale student behavior dataset \citep{zhou2023stuart} (AB) & hand-raising(70k), standing(21k), sleeping(3k), yawning(3,216) and smiling(129k), techear(15k), 36k images \\
    \midrule
    Classroom behavior dataset \citep{zhao2023bitnet} & writing, reading, listening, raising hand, turning around, standing, discussing, and guiding, 4432 images and 151574 annotation boxes\\
    \midrule
    \cite{qin2024improved} & eating(1,200), raising hands(1,000), reading(1,000), sleeping on the desk(1,000), and writing(1,000), 5200 images\\
    \midrule
    DBS Dataset~\cite{liu2024improved} & listening, raising hands, standing up, reading, writing, looking around, lying on the desk, discussing, and other behaviors, with a total of 6890 annotated images.\\
   \bottomrule
\end{tabular}
\end{table*}


\subsection{SCB-Dataset Statistics and Training results}

Table~\ref{SCB-Dataset-train-val-statistics} shows the SCB data statistics. Table~\ref{Object-Detection-SCB-Dataset-YOLOv5}, Table~\ref{Object-Detection-SCB-Dataset-YOLOv7-teacher-behavior}, Table~\ref{Object-Detection-SCB-Dataset-YOLOv8}, Table~\ref{Object-Detection-SCB-Dataset-YOLOv9}, Table~\ref{Object-Detection-SCB-Dataset-YOLOv10}, Tabel~\ref{Object-Detection-SCB-Dataset-YOLOv11}, Table~\ref{Object-Detection-SCB-Dataset-YOLOv12}, and Table~\ref{Object-Detection-SCB-Dataset-YOLOv13} show the training results of SCB on YOLOv5, v7, v8, v9, v10, v11, v12 and v13.

\begin{table}[ht]

\caption{SCB-Dataset training and validation data statistics}
\label{SCB-Dataset-train-val-statistics}

\small
\centering
\renewcommand{\arraystretch}{1.4}
\begin{tabular}{l|c|c}

    \toprule
    Class: Object Detection & Train & Val \\
    \midrule
    hand-raising & 10538 & 2915 \\
    read & 17539 & 6539 \\
    write & 6447 & 3394 \\
    discuss & 3607 & 1785 \\
    bow the head & 4422 & 540 \\
    turn the head & 7943 & 3213 \\
    guide & 1155 & 449 \\
    answer & 2574 & 853 \\
    on-stage interaction & 528 & 149 \\
    blackboard-writing & 821 & 277 \\
    teacher & 8490 & 3228 \\
    stand & 13932 & 4967 \\
    screen & 5025 & 1959 \\
    blackboard & 7847 & 3445 \\
    \midrule
    Class: Image Classification & Train & Val \\
    \midrule
    hand-raising & 1472 & 187 \\
    read and write & 814 & 98 \\
    discuss & 703 & 51 \\
    student blackboard-writing & 211 & 17 \\
    on-stage presentation & 34 & 10 \\
    answering questions & 639 & 52 \\
    reading aloud & 134 & 13\\
    listen & 2294 & 157 \\
    guide & 1584 & 185 \\
    answer & 3938 & 439 \\
    on-stage interaction & 816 & 113 \\
    blackboard-writing & 1703 & 204 \\
    teach & 3088 & 240 \\
    patrol & 1722 & 101 \\
\bottomrule
\end{tabular}

\end{table}

\begin{table}[ht]
\setlength{\tabcolsep}{1mm}
\fontsize{9pt}{11pt}\selectfont
\centering
\renewcommand{\arraystretch}{1.4}
\caption{The training results of Object Detection Dataset in SCB-Dataset (teacher behavior part) on YOLOv5.}
\label{Object-Detection-SCB-Dataset-YOLOv5}
\begin{tabular}{lcccc}

    \toprule
    \textbf{class} & P & R & mAP@0.5 & mAP@.95 \\
    \midrule
    all & 81.8 & 84.1 & 88.1 & 67.3  \\
    guide & 68.2 & 59.1 & 66.0 & 31.5  \\
    answer & 67.4 & 81.7 & 80.8 & 57.1 \\
    On-stage interaction & 52.0 & 82.8 & 80.0 & 59.2  \\
    blackboard-writing & 97.7 & 70.4 & 96.3 & 73.9 \\
    teacher & 95.1 & 91.9 & 95.7 & 71.9 \\
    stand & 88.9 & 92.3 & 90.4 & 63.9 \\
    screen & 91.6 & 97.5 & 98.0 & 90.0 \\
    blackBoard & 93.4 & 97.1 & 98.0 & 90.7 \\
\bottomrule
\end{tabular}
\end{table}



\begin{table}[htbp]
\caption{The training results of SCB-Dataset on YOLOv7.}
\label{SCB-Dataset-YOLOv7-train}
\setlength{\tabcolsep}{1mm}
\fontsize{9pt}{11pt}\selectfont
\begin{center}
\begin{tabular}{l|l|l|l|l|l}
   \toprule
   Dataset& class & P & R & mAP@0.5 & mAP@.95 \\
   \midrule
   \multirow{4}{*}{\makecell[l]{SCB5-A}} & all & 71.1 & 70.9 & 74.0 & 56.8  \\
    & hand-raising & 79.4 & 76.9 & 79.2 & 59.4   \\
    & read & 65.5 & 68.2 & 70.5 & 52.9 \\
    & write & 68.4 & 67.8 & 72.2 & 58.1 \\
    \midrule
  \multirow{3}{*}{\makecell[l]{SCB5-B}} & all & 94.5 & 97.3 & 98.7 & 91.9  \\
    & screen & 94.8 & 95.7 & 98.2 & 95.1   \\
    & backboard & 94.2 & 98.9 & 99.2 & 88.8 \\
    \midrule
   \multirow{1}{*}{\makecell[l]{SCB5-C}} & all/discuss & 67.5 & 72.5 & 74.7 & 39.3  \\
   
    \midrule
   \multirow{4}{*}{\makecell[l]{SCB5-D}} & all & 85.5 & 82.6 & 86.4 & 67.2  \\
    & guide & 88.0 & 81.7 & 87.0 & 49.5   \\
    & answer & 89.3 & 88.0 & 92.3 & 76.7 \\
    & stage interaction & 69.9 & 65.2 & 68.5 & 54.7 \\
    & board writing & 94.5 & 95.6 & 97.7 & 87.9 \\
    \midrule
   \multirow{1}{*}{\makecell[l]{SCB5-E}} & all/stand & 95.8 & 91.7 & 96.6 & 80.5  \\
    \midrule
   \multirow{1}{*}{\makecell[l]{SCB5-F}} & all/teacher & 96.2 & 94.4 & 97.7 & 82.7  \\
    \midrule
   \multirow{3}{*}{\makecell[l]{SCB5-G}} & all &  &  &  &   \\
    & bow-head & - & - & - & -   \\
    & turn-head & - & - & - & - \\
    \midrule
   \multirow{1}{*}{\makecell[l]{SCB5-H}} & all/talk & 87.8 & 62.6 & 77.2 & 61.3  \\
   \bottomrule
\end{tabular}
\end{center}
\end{table}

\begin{table}[ht]

\setlength{\tabcolsep}{1mm}
\fontsize{9pt}{11pt}\selectfont
\centering
\renewcommand{\arraystretch}{1.4}
\begin{tabular}{lcccc}

    \toprule
    \textbf{class} & P & R & mAP@0.5 & mAP@.95 \\
    \midrule
    all & 91.1 & 90.9 & 94.0 & 80.8  \\
    guide & 88.5 & 78.3 & 83.6 & 48.9  \\
    answer & 86.2 & 86.6 & 91.5 & 80.8\\
    On-stage interaction & 82.3 & 84.5 & 90.1 & 81.5  \\
    blackboard-writing & 91.0 & 93.5 & 96.4 & 86.6 \\
    teacher & 95.5 & 95.2 & 97.7 & 83.0 \\
    stand & 93.1 & 94.7 & 96.6 & 79.8 \\
    screen & 96.1 & 97.1 & 97.9 & 92.5 \\
    blackBoard & 96.2 & 97.1 & 98.1 & 93.3 \\
\bottomrule
\end{tabular}
\caption{The training results of Object Detection Dataset in SCB-Dataset (teacher behavior part) on YOLOv7.}
\label{Object-Detection-SCB-Dataset-YOLOv7-teacher-behavior}
\end{table}

\begin{table}[ht]

\setlength{\tabcolsep}{1mm}
\fontsize{9pt}{11pt}\selectfont
\centering
\renewcommand{\arraystretch}{1.4}
\caption{The training results of Object Detection Dataset in SCB-Dataset (teacher behavior part) on YOLOv8.}
\label{Object-Detection-SCB-Dataset-YOLOv8}
\begin{tabular}{lcccc}

    \toprule
    \textbf{class} & P & R & mAP@0.5 & mAP@.95 \\
    \midrule
    all & 90.6 & 89.2 & 93.6 & 83.1  \\
    guide & 81.5 & 67.5 & 79.4 & 53.5  \\
    answer & 87.2 & 87.5 & 92.9 & 86.4\\
    On-stage interaction & 82.5 & 83.2 & 88.9 & 80.1  \\
    blackboard-writing & 90.8 & 93.9 & 97.6 & 87.3 \\
    teacher & 96.2 & 93.8 & 97.3 & 86.0 \\
    stand & 94.0 & 93.7 & 96.5 & 83.0 \\
    screen & 95.8 & 96.7 & 97.9 & 93.4 \\
    blackBoard & 96.5 & 97.2 & 98.3 & 94.7 \\
\bottomrule
\end{tabular}
\end{table}

\begin{table}[ht]

\setlength{\tabcolsep}{1mm}
\fontsize{9pt}{11pt}\selectfont
\centering
\renewcommand{\arraystretch}{1.4}
\caption{The training results of Object Detection Dataset in SCB-Dataset (teacher behavior part) on YOLOv9.}
\label{Object-Detection-SCB-Dataset-YOLOv9}
\begin{tabular}{lcccc}

    \toprule
    \textbf{class} & P & R & mAP@0.5 & mAP@.95 \\
    \midrule
    all & 87.8 & 87.7 & 91.8 & 78.2  \\
    guide & 79.3 & 63.7 & 74.3 & 45.0  \\
    answer & 79.7 & 84.5 & 89.5 & 78.7\\
    On-stage interaction & 71.7 & 79.9 & 84.7 & 72.6  \\
    blackboard-writing & 95.0 & 95.1 & 97.9 & 84.5 \\
    teacher & 94.2 & 93.1 & 96.9 & 81.8 \\
    stand & 92.2 & 92.5 & 95.4 & 77.1 \\
    screen & 94.6 & 96.3 & 97.6 & 90.9 \\
    blackBoard & 95.5 & 96.6 & 98.2 & 94.4 \\
\bottomrule
\end{tabular}
\end{table}

\begin{table}[ht]
\setlength{\tabcolsep}{1mm}
\fontsize{9pt}{11pt}\selectfont
\centering
\renewcommand{\arraystretch}{1.4}
\caption{The training results of Object Detection Dataset in SCB-Dataset (teacher behavior part) on YOLOv10.}
\label{Object-Detection-SCB-Dataset-YOLOv10}
\begin{tabular}{lcccc}

    \toprule
    \textbf{class} & P & R & mAP@0.5 & mAP@.95 \\
    \midrule
    all & 86.9 & 84.1 & 90.1 & 76.2  \\
    guide & 79.7 & 58.3 & 71.1 & 41.8  \\
    answer & 80.9 & 78.4 & 86.1 & 77.0 \\
    On-stage interaction & 80.1 & 78.6 & 86.0 & 77.5  \\
    blackboard-writing & 86.3 & 88.6 & 94.7 & 79.2 \\
    teacher & 90.1 & 88.7 & 95.1 & 77.1 \\
    stand & 88.1 & 88.5 & 93.1 & 72.6 \\
    screen & 94.9 & 96.0 & 97.4 & 90.9 \\
    blackBoard & 95.2 & 95.6 & 97.6 & 93.5 \\
\bottomrule
\end{tabular}
\end{table}

\begin{table}[ht]
\centering

\setlength{\tabcolsep}{1mm}
\fontsize{9pt}{11pt}\selectfont
\renewcommand{\arraystretch}{1.4}
\caption{The training results of Object Detection Dataset in SCB-Dataset (teacher behavior part) on YOLOv11.}
\label{Object-Detection-SCB-Dataset-YOLOv11}
\begin{tabular}{lcccc}

    \toprule
    \textbf{class} & P & R & mAP@0.5 & mAP@.95 \\
    \midrule
    all & 87.6 & 90.1 & 92.9 & 81.8  \\
    guide & 72.2 & 66.7 & 72.1 & 45.9  \\
    answer & 87.1 & 88.8 & 94.1 & 86.6 \\
    On-stage interaction & 75.1 & 89.9 & 91.2 & 82.4  \\
    blackboard-writing & 92 & 93.4 & 97.3 & 86.5 \\
    teacher & 94.8 & 94.1 & 97.0 & 85.0 \\
    stand & 91.9 & 93.5 & 95.8 & 81.5 \\
    screen & 92.7 & 96.9 & 97.4 & 92.1 \\
    blackBoard & 94.9 & 97.3 & 98.3 & 93.9 \\
\bottomrule
\end{tabular}
\end{table}

\begin{table}[ht]

\setlength{\tabcolsep}{1mm}
\fontsize{9pt}{11pt}\selectfont
\centering
\renewcommand{\arraystretch}{1.4}
\caption{The training results of Object Detection Dataset in SCB-Dataset (teacher behavior part) on YOLOv12.}
\label{Object-Detection-SCB-Dataset-YOLOv12}
\begin{tabular}{lcccc}

    \toprule
    \textbf{class} & P & R & mAP@0.5 & mAP@.95 \\
    \midrule
    all & 86.2 & 86.6 & 90.6 & 77.1  \\
    guide & 78.6 & 58.9 & 71.8 & 43.7  \\
    answer & 77.8 & 83.1 & 88.0 & 78.9 \\
    On-stage interaction & 66.5 & 81.2 & 82.6 & 69.4  \\
    blackboard-writing & 91.4 & 93.5 & 96.4 & 84.2 \\
    teacher & 92.6 & 91.9 & 95.9 & 80.2 \\
    stand & 91.4 & 90.8 & 94.5 & 75.6 \\
    screen & 95.6 & 96.4 & 97.7 & 92.1 \\
    blackBoard & 95.7 & 96.7 & 98.1 & 92.9 \\
\bottomrule
\end{tabular}
\end{table}

\begin{table}[ht]
\setlength{\tabcolsep}{1mm}
\fontsize{9pt}{11pt}\selectfont
\centering
\renewcommand{\arraystretch}{1.4}
\caption{The training results of Object Detection Dataset in SCB-Dataset (teacher behavior part) on YOLOv13.}
\label{Object-Detection-SCB-Dataset-YOLOv13}
\begin{tabular}{lccccc}

    \toprule
    \textbf{class} & P & R & mAP@0.5 & mAP@0.75 & mAP@.95 \\
    \midrule
    all & 89.6 & 85.1 & 91.4 & 82.6 & 77.2 \\
    guide & 84.6 & 54.6 & 74.8 & 45.1 & 43.4 \\
    answer & 82.9 & 80.6 & 87.6 & 80.4 & 76.0\\
    On-stage interaction & 79.0 & 78.2 & 87.1 & 83.5 & 79.8 \\
    blackboard-writing & 93.3 & 92.7 & 96.1 & 90.2 & 81.7 \\
    teacher & 93.5 & 91.7 & 95.8 & 86.5 & 78.0 \\
    stand & 92.2 & 89.0 & 94.0 & 82.3 & 73.7 \\
    screen & 95.2 & 97.0 & 98.0 & 96.7 & 92.6 \\
    blackBoard & 95.9 & 96.7 & 98.1 & 96.1 & 92.9 \\
\bottomrule
\end{tabular}
\end{table}

\vspace{12pt}

\subsection{Prompt}

Student behaviors and teacher behaviors are divided into two categories.
Student behaviors include: read and write, on-stage presentation, student blackboard-writing, answering questions, reading aloud, discuss, listen, hand-raising, others.
Teacher behaviors include: teach, guide, answer, on-stage interaction, blackboard-writing, patrol, others.

The definitions of each category are as follows:
Definitions of student behavior categories:
Read and write: Students are reading or writing.
On-stage presentation: Students are presenting on stage. The difference from the on-stage interaction in teacher behaviors is that on-stage interaction involves teachers, while on-stage presentation involves only students on stage without teachers.
Student blackboard-writing: Students are writing on the blackboard. Note the difference between student blackboard-writing and teacher blackboard-writing.
Answering questions: Students stand up to answer questions. Note the difference from the answer in teacher behaviors. Answering questions means there are only students in the picture without teachers, while answer means there are both students and teachers in the picture.
Reading aloud: Students read aloud in unison. Note the difference from read and write. When students read aloud, they open their mouths or have a tendency to open their mouths on the basis of reading and writing.
Discuss: Students discuss in class, which can be a discussion between deskmates or between students in the front and back rows.
Listen: Students look up to listen to the teacher's lecture.
Hand-raising: Students raise their hands. Generally, hand-raising is counted only when more than 3 students raise their hands.
Others: Any behavior that does not belong to the above categories.

Definitions of teacher behavior categories:
Teach: Teachers usually stand on the podium and explain knowledge points in class. Note the difference between teacher-student interaction and teacher teaching. In teacher teaching, only the teacher is standing.
Guide: Teachers step down from the podium to give individual guidance to a certain student, usually accompanied by actions such as bending over and stopping (just standing next to the student to watch is not considered as guide).
Answer: Students answer the teacher's questions. Usually, both the teacher and the student are standing, with the teacher asking questions and the student answering. Note the difference between teacher teaching and teacher-student interaction. In teacher teaching, no student stands up to answer questions.
On-stage interaction: Teachers invite students to the stage for activities, including playing games, completing tasks or students writing on the blackboard on stage. Note the difference between the on-stage presentation in student behaviors and the on-stage interaction in teacher behaviors. On-stage presentation means there are only students on the podium, while on-stage interaction means there are both teachers and students.
Blackboard-writing: Teachers write on the blackboard. Note that blackboard-writing refers to the teacher's writing behavior on the blackboard, and students' writing on the stage is not considered as the teacher's behavior.
Patrol: Teachers are not on the podium but walk around the classroom, observing students or patrolling the classroom.
Others: Any behavior that does not belong to the above categories.

Recognition rules:

1. Priority of single behavior: Only one dominant behavior is identified for each picture. If there are compound actions, classify according to the dominant behavior.
2. Unique output: Only one behavior category is output for each recognition.

Now you need to identify the students' behaviors in the picture. Output format:
Please output the behavior category strictly according to the following format:
read and write/on-stage presentation/student blackboard-writing/answering questions/reading aloud/discuss/hand-raising/listen/others

\subsection{Behavior Description}

Finally, the description of behavior classes is as follows:

\begin{tcolorbox}[
    colframe=black,
    colback=gray!10,
    arc=1mm,
    boxrule=0.3mm,
    fontupper=\fontsize{9pt}{11pt}\selectfont,
    boxsep=-1pt,           
]
\textbf{1.hand-raising}: Students raise their hands in class to indicate they want to speak or ask a question. 

\textbf{2.read}: Students read books, textbooks, or notes in class.

\textbf{3.write}: Students take notes or complete written assignments in class.

\textbf{4.discuss}: Students discuss classroom content with each other. 

\textbf{5.bow the head}: Students lower their heads to look at the desk or items in their hands, possibly being distracted or focused on personal activities.

\end{tcolorbox}

\begin{tcolorbox}[
    colframe=black,
    colback=gray!10,
    arc=1mm,
    boxrule=0.3mm,
    fontupper=\fontsize{9pt}{11pt}\selectfont,
    boxsep=-1pt,           
]
\textbf{6.turn the head}: Students turn their heads, possibly to look at classmates/teachers or events happening in the classroom. 

\textbf{7.blackboard-writing}: Teachers/students write on the blackboard or draw. 

\textbf{8.on-stage presentation}: Students stand up and answer questions. The difference between "answering questions" and "answer" is that "answering questions" means there are only students in the image, no teachers, while "answer" means there are both students and teachers in the image.

\textbf{9.reading aloud}: It is generally students reading aloud the content in books.

\textbf{10.listen}: Students look up to listen to the teacher's lecture.

\end{tcolorbox}

\begin{tcolorbox}[
    colframe=black,
    colback=gray!10,
    arc=1mm,
    boxrule=0.3mm,
    fontupper=\fontsize{9pt}{11pt}\selectfont,
    boxsep=-1pt,           
]
\textbf{11.guide}: Teachers provide guidance or explanations to students in class.

\textbf{12.answer}: Students respond to the teacher’s questions or instructions. 

\textbf{13.on-stage interaction}: Students interact with teachers or other students on the stage.

\textbf{14.teacher}: Teacher identity, used to distinguish between students and teachers, functions such as locating the teacher's coordinates.

\textbf{15.teach}: It is generally teachers standing on the podium, explaining knowledge points.
\end{tcolorbox}

\begin{tcolorbox}[
    colframe=black,
    colback=gray!10,
    arc=1mm,
    boxrule=0.3mm,
    fontupper=\fontsize{9pt}{11pt}\selectfont,
    boxsep=-1pt,           
]
\textbf{16.patrol}: Teachers walk around the classroom, observing students or patrolling the classroom.

\textbf{17.stand}: Students or teachers stand in class.

\textbf{18.screen}: The screen where teachers show PPTs

\textbf{19.blackboard}: The blackboard where teachers write on the blackboard.

\end{tcolorbox}


\end{document}